\pdfoutput=1

\documentclass[11pt]{article}

\usepackage{acl}
\usepackage{amsmath}
\usepackage{times}
\usepackage{latexsym}
\usepackage{arydshln}
\usepackage{booktabs}
\usepackage{tikz}
\usepackage[edges]{forest}
\usepackage{pifont}
\usepackage{amssymb}
\usepackage{makecell}
\usepackage{adjustbox}
\usepackage{multirow}
\newcommand{\cmark}{\textcolor{DarkGreen}{\ding{51}}}
\newcommand{\xmark}{\textcolor{red}{\ding{55}}}%

\newcommand{\ie}{\emph{i.e.}}

\definecolor{lightergray}{RGB}{230,230,230}
\definecolor{DarkGreen}{RGB}{30,130,30}
\usepackage[T1]{fontenc}

\usepackage[utf8]{inputenc}

\usepackage{microtype}

\title{Unveiling the Spectrum of Data Contamination in Language Models: \\ A Survey from Detection to Remediation}
\author{Chunyuan Deng$^{1,2}\thanks{~~Equal Contribution. }$ \quad Yilun Zhao$^{1*}$ \quad Yuzhao Heng$^2$ \quad Yitong Li$^2$ \\
\bf{\quad Jiannan Cao$^3$\quad Xiangru Tang$^1$ \quad Arman Cohan$^{1,4}$ \vspace{4pt}}\\
$^1$Yale University \quad $^2$Georgia Institute of Technology 
\quad $^3$MIT \quad $^4$Allen Institute for AI\\
\texttt{\{cd2249,yilun.zhao,arman.cohan\}@yale.edu} 
}

\begin{document}
\maketitle
\begin{abstract}
Data contamination has garnered increased attention in the era of large language models (LLMs) due to the reliance on extensive internet-derived training corpora. The issue of training corpus overlap with evaluation benchmarks—referred to as contamination—has been the focus of significant recent research. This body of work aims to identify contamination, understand its impacts, and explore mitigation strategies from diverse perspectives. However, comprehensive studies that provide a clear pathway from foundational concepts to advanced insights are lacking in this nascent field. Therefore, we present a comprehensive survey in the field of data contamination, 
laying out the key issues, methodologies, and findings to date, and highlighting areas in need of further research and development. In particular, we begin by examining the effects of data contamination across various stages and forms. We then provide a detailed analysis of current contamination detection methods, categorizing them to highlight their focus, assumptions, strengths, and limitations. We also discuss mitigation strategies, offering a clear guide for future research. This survey serves as a succinct overview of the most recent advancements in data contamination research, providing a straightforward guide for the benefit of future research endeavors.\footnote{List of relevant papers/resources to the survey will be also actively maintained \url{https://github.com/yale-nlp/lm-contamination-survey}.}
\end{abstract}

\section{Introduction}
Data contamination refers to \textit{the accidental or deliberate inclusion of evaluation or benchmark data in the training phase of language models}, resulting in artificially high benchmark scores~\cite{schaeffer2023pretraining}. This issue, while longstanding—stemming from the foundational ML principle of separating training and test sets—has garnered increased attention with the advent of large language models (LLMs). These models are trained on vast corpora sourced from the web~\cite{openai2023gpt4, touvron2023llama}, heightening the risk that training data may inadvertently encompass instances from evaluation benchmarks~\cite{brown2020language,chowdhery2022palm,touvron2023llama,touvron2023llama2}. Such contamination of evaluation benchmarks can obscure the true generalization performance of LLMs, as it might artificially inflate benchmark scores by testing the models' ability to ``memorize'' and ``recall'' rather than ``reason'' or ``generalize''. 

\begin{figure}[t]
    \centering
    \includegraphics[width=0.50\textwidth]{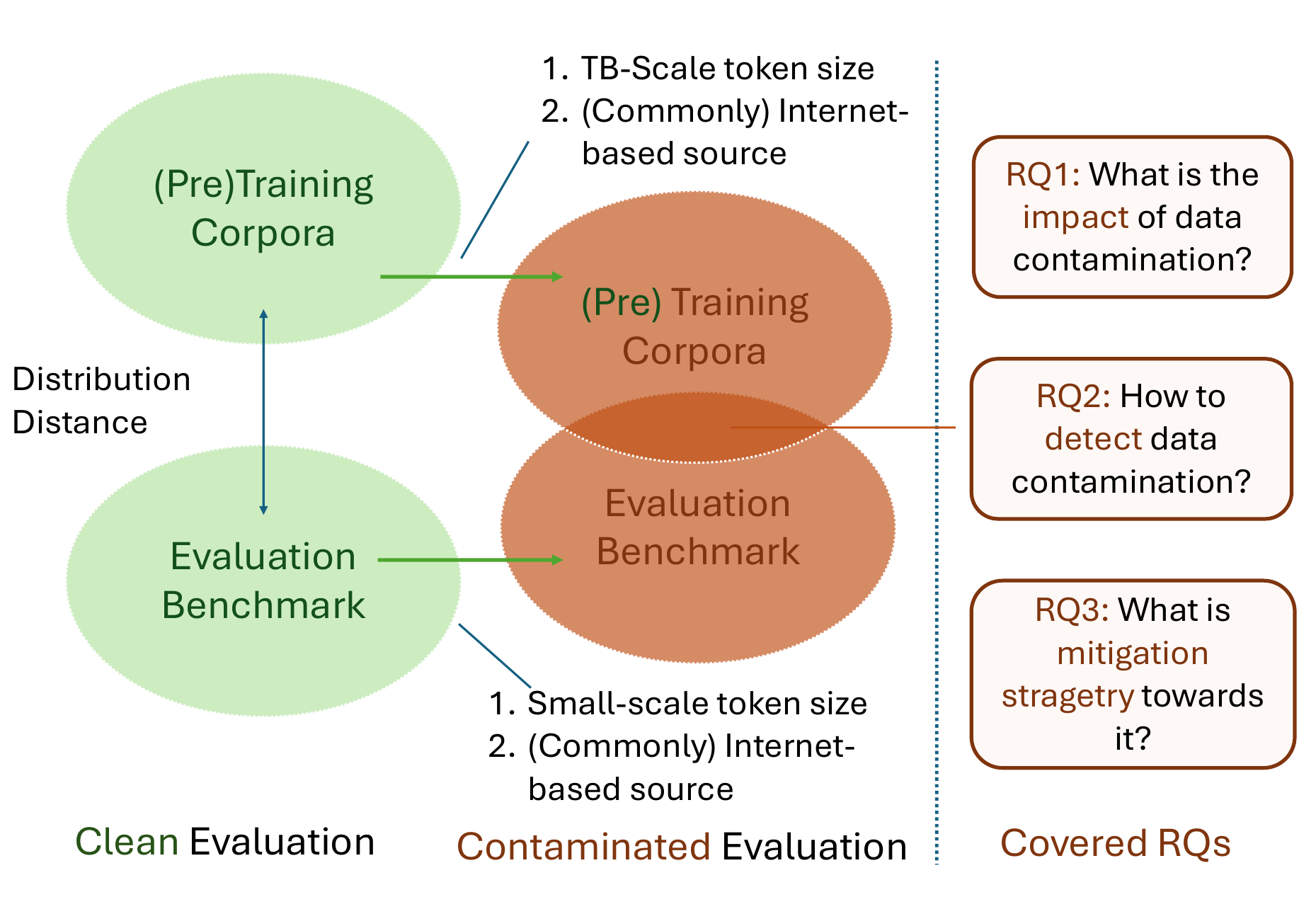} 
    \caption{Basic illustration of data contamination and the research questions related to it. Clean evaluation is defined as having no overlap between the pre-training corpora and the benchmarks, and contaminated evaluation is defined as having a significant overlap between them.} 
    \label{fig:contamination_intro}
        \vspace{-.4cm}
\end{figure}

Given the increasing concerns about potential contamination of evaluation benchmarks and their broader impact on downstream task performance, numerous recent studies have focused on identifying and mitigating data contamination in these benchmarks. These efforts aim to better understand how contamination affects our perception of model capabilities. In general, research on data contamination could be broadly categorized into two main areas: (i) investigations of models trained with open-source data and (ii) studies relevant to models developed using proprietary data. Generally, having access to training data, or the lack thereof, has a profound influence on modern contamination research.

\definecolor{mycolor}{RGB}{215, 245, 200}
\definecolor{hidden-draw}{RGB}{20,68,106}

\tikzstyle{my-box}=[
    rectangle,
    draw=hidden-draw,
    rounded corners,
    text opacity=1,
    minimum height=1.5em,
    minimum width=5em,
    inner sep=2pt,
    align=center,
    fill opacity=.5,
    line width=0.8pt,
]
\tikzset{
leaf/.style={
my-box,
minimum height=1.5em,
fill={brown}, %
text=black,
align=center,
font=\footnotesize,
inner xsep=2pt,
inner ysep=4pt,
line width=0.8pt
}
}
\begin{figure*}[!t]
    \centering
    \resizebox{\textwidth}{!}{
        \begin{forest}
            forked edges,
            for tree={
                grow=east,
                reversed=true,
                anchor=base west,
                parent anchor=east,
                child anchor=west,
                base=center,
                font=\large,
                rectangle,
                draw=brown,
                rounded corners,
                align=left,
                text centered,
                minimum width=5em,
                edge+={darkgray, line width=1.3pt},
                s sep=6pt,
                inner xsep=2pt,
                inner ysep=5pt,
                line width=0.8pt,
                ver/.style={rotate=90, child anchor=north, parent anchor=south, anchor=center},
            },
            where level=1{text width=11em,font=\normalsize,}{},
            where level=2{text width=11em,font=\normalsize,}{},
            where level=3{text width=11em,font=\normalsize,}{},
            where level=4{text width=7em,font=\normalsize,}{},
            [
                Data Contamination
                [
                    Task
                    [
                        Definition 
                    ]
                    [
                        Urgency 
                    ]
                    [
                        Domain 
                        [
                            White-box \\Language Models
                            [
                                Bert~\cite{devlin2019bert}{,} GPT~\cite{brown2020language}{,}\\OLMo~\cite{groeneveld2024olmo}{,}Llama \cite{touvron2023llama}
                                , leaf, text width=21em
                            ]
                        ]
                        [
                            Gray-box \\Large Language Models
                            [
                                Mistral~\cite{jiang2023mistral}{, }Qwen~\cite{bai2023qwen}{,}\\Falcon~\cite{mei2022falcon}{, }etc
                                , leaf, text width=21em
                            ]
                        ]
                        [
                            Black-box \\Large Language Models
                            [
                                ChatGPT~\cite{chatgpt}{, } GPT-4~\cite{openai2023gpt4}{, }\\ Gemini~\cite{Anil2023GeminiAF}{,} Claude~\cite{claude}{,}etc.\\
                                 , leaf, text width=21em
                            ]
                        ]
                    ]
                ]
                [
                    Impact of Contamination
                        [
                            ~\citet{geva2021transformer}{,}~\citet{magar2022data}{,}~\citet{Blevins2022LanguageCH}{, }\\{,}\citet{hartmann2023sok}{,}~\citet{jiang2024investigating}{, }~\citet{zhu2024critical}{,}~\citet{duan2024membership}\\\citet{geva2023dissecting}{,}\citet{haviv2023understanding}{,}\citet{srivastava2023imitation}
                            , leaf, text width=28em
                        ]                
                ]
                [
                    Detection
                    [
                        Retrieval     
                        [
                        Model Developer-Side
                        [
                            GPT-3~\cite{brown2020language}{,} PaLM~\cite{chowdhery2022palm}{,}\\
                            Llama~\cite{touvron2023llama}, leaf, text width=22em 
                        ]
                        ]
                        [
                        Academic Community-Side
                        [
                            ~\citet{dodge2021documenting}{, }~\citet{piktus2023roots}{, }~\citet{elazar2023whats}\\~\citet{kandpal2023large}{, }~\citet{deng2023investigating}{,}~\citet{riddell2024quantifying}\\~\citet{balloccu2024leak}, leaf, text width=22em
                        ]
                        ]
                    ]
                    [
                        Temporal Cutoff
                        [
                            Pretrain-Level 
                            [
                                ~\citet{shi2023detecting}, leaf, text width=8em
                            ]
                            Task-Level 
                            [
                                ~\citet{li2023task}{, }~\citet{roberts2023data}{, }\\~\citet{aiyappa2023trust}, leaf, text width=18em
                            ]
                        ]
                    ]
                    [
                        Masking-based
                        [
                            Book-Level
                            [
                                ~\citet{chang2023speak}, leaf, text width=8em
                            ]
                        ]
                         [
                            Benchmark-Level 
                            [
                                ~\citet{deng2023investigating}{,}~\citet{bordt2024elephants}{,}~\citet{xu2024benchmarking}, leaf, text width=22em
                            ]
                        ]
                    ]
                    [
                        Perturbation-based
                            [
                                ~\citet{Wei2023SkyworkAM}{, }~\citet{yang2023rethinking}{,}~\citet{dekoninck2024constat}\\~\citet{dekoninck2024constat}{,}~\citet{ranaldi2024investigating}, leaf, text width=22em
                            ]  
                    ]
                    [
                        Canonical Order
                            [
                                ~\citet{oren2023proving}, leaf, text width=18em
                            ]
                    ]
                    [
                        Behavior Manipulation
                            [
                                ~\citet{golchin2023time}{,}~\citet{Golchin2023DataCQ}\\~\citet{dong2024generalization}, leaf, text width=22em
                            ]
                    ]
                    [
                        Membership Inference\\Attacks
                            [
                                ~\citet{yeom2018privacy}{,}~\citet{carlini2021extracting}{,}~\citet{carlini2022membership}\\{,}~\citet{mattern2023membership}{,}~\citet{shi2023detecting}{,}~\citet{xu2024benchmarking}\\~\citet{ye2024data}, leaf, text width=22em
                            ]
                    ]
                ]
                [
                    Mitigation
                    [
                        Evaluation 
                            [
                                ~\citet{Zhu2023DyValGD}{,}~\citet{zhu2023cleaneval}{, }~\citet{li2023latesteval}, leaf, text width=22em
                            ]
                    ]
                    [
                        Guideline
                            [
                                ~\citet{Jacovi2023StopUT}{,}~\citet{Zhou2023DontMY}{, }~\citet{Sainz2023NLPEI}, leaf, text width=22em
                            ]
                    ]         
                ]
            ]
    \end{forest}}
\vspace{-4mm}
\caption{Taxonomy of research on Data Contamination in large language models that consists of the task, effect, detection and mitigation.}
    \label{fig:tree}
\end{figure*}
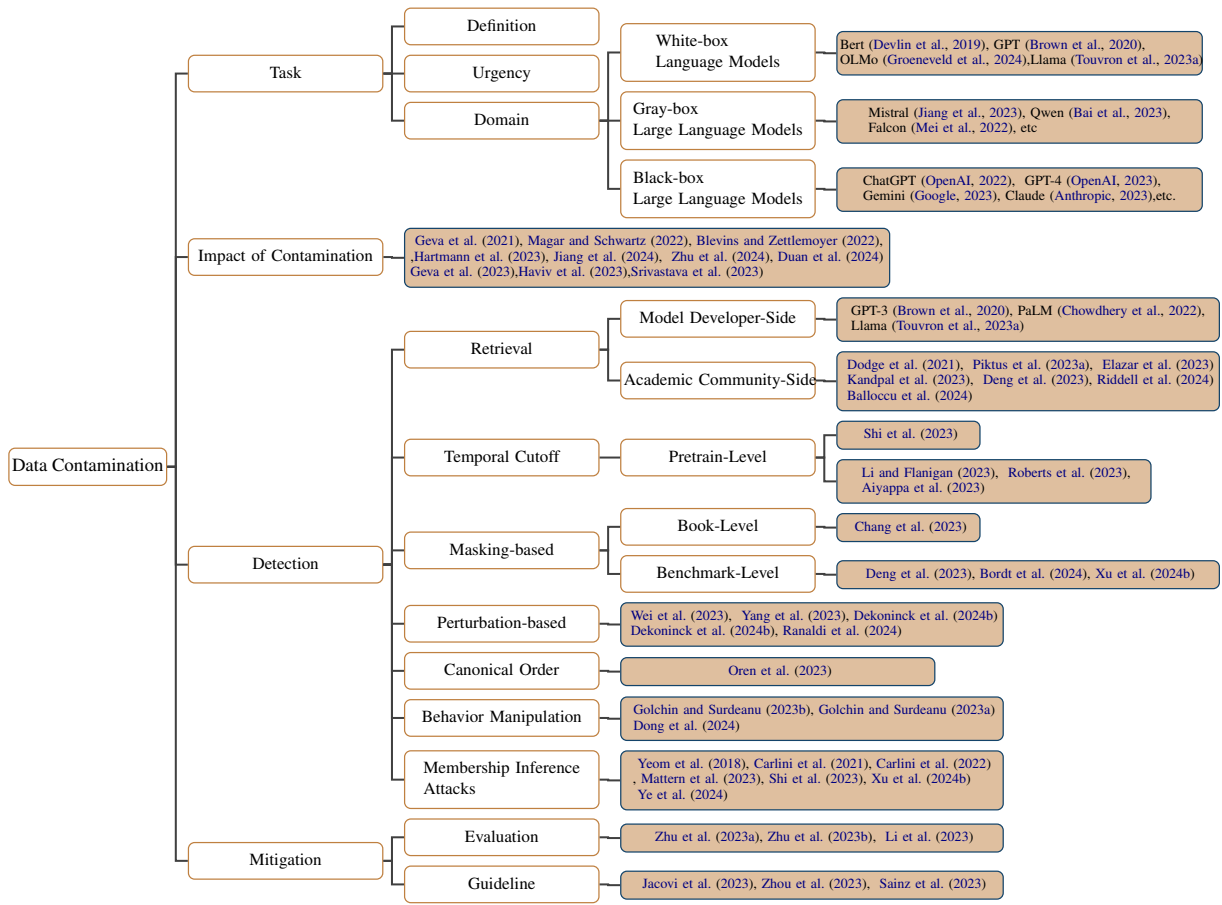

In this paper, we present a comprehensive analysis of the growing field of data contamination detection and mitigation. Our objective is to delve into the downstream impacts of data contamination, investigate existing methods for detecting data contamination, and discuss a range of mitigation strategies.
The paper is structured as outlined in Figure~\ref{tab:dataset_comparison}. We start by establishing the background of data contamination (\S \ref{background}) and discussing the effect of contamination (\S \ref{mem_effect}). Following this, We provide a detailed analysis of current methods for detecting data contamination (\S \ref{detect}). We categorize these methods and critically examine the assumptions each relies on, highlighting the prerequisites and limitations for their application. Subsequently, we explore strategies for mitigating data contamination (\S \ref{mitigating}), tackling potential hurdles, and proposing avenues for future investigations in this domain. Together with concurrent studies on data contamination~\cite{ravaut2024llms,xu2024benchmark}, this paper aims to furnish NLP researchers with an in-depth, systematic understanding of data contamination issues, thereby making a significant contribution to enhancing the integrity of evaluations in the field.

\section{Background} \label{background}

To provide a comprehensive understanding of data contamination, this section delves into its definition, the urgency of addressing it, and its implications across different types of language models. 
\paragraph{What is data contamination?}
Data contamination occurs when benchmark or test set data are inadvertently included in the training phase. This issue is particularly relevant when evaluating LLMs that have been partially trained with a test set from a benchmark, potentially leading to an inflated performance score. This phenomenon, known as data contamination,  is critical for ensuring fairness and unbiased evaluation in modern LLMs. 

\paragraph{Significance of studying contamination.}
Thorough and complete evaluation of LLM capabilities has remained a largely unsolved problem, with benchmark contamination playing a critical role in achieving a comprehensive assessment of LLM capabilities. In traditional NLP and ML, it is often rather straightforward to separate training and testing data, allowing for evaluating models' generalization capabilities to new and unseen cases \citep{suhr-etal-2020-exploring, talmor-berant-2019-multiqa, pmlr-v80-lake18a}. However, with web-scale training data of LLMs and their enormous size in terms of a number of parameters, such clear separation has become very difficult. Especially because many existing NLP benchmarks are already constructed from web data (e.g., news articles, Wikipedia data, scientific papers, social media, etc), and new benchmarks are publicly released on the internet as well, making it possible for them to be included in future training sets through various data collection means. Thus, contamination of evaluation benchmarks has led to an incomplete understanding at best and, at worst, a misleading assessment of the true capabilities of LLMs.

\paragraph{Language model types in data contamination.\\} 
(1) \emph{White-box Language Models}: The white-box language model refers to the model whose internal workings, such as the model architecture, parameters, and training data, are transparent, allowing for a deeper understanding and analysis of its behavior. In the realm of data contamination, the focus often centers on models like BERT~\cite{devlin2019bert} and GPT-2~\cite{radford2019language}, or larger models like Pythia~\cite{biderman2023pythia} or OLMo~\cite{groeneveld2024olmo}, to examine the \emph{impacts of contamination} (\S \ref{mem_effect}). This involves exploring the correlation between the contaminated data and downstream task performance from the perspective of how well these models memorize and are influenced by the contaminated input. 

\noindent (2) \emph{Gray-box Language Models}: 
The gray-box language model is a type of language model that provides some level of transparency and interpretability into its internal workings, such as revealing certain architectural components or allowing limited access to its training data, while still maintaining a degree of opacity or abstraction over other aspects of the model. This typically refer to large-scale models, such as LLaMA~\cite{touvron2023llama, touvron2023llama2}, Mistral~\cite{jiang2023mistral}, Qwen~\cite{bai2023qwen}, and Phi-3~\cite{abdin2024phi3}. Although the extent of openness varies among these models, they are generally characterized by their accessibility. This accessibility facilitates extensive research into their architectures and training datasets, enabling the development and validation of innovative methodologies within the field.

\noindent(3) \emph{Black-box Language Models}: Black-box LLMs often refer to proprietary models such as ChatGPT~\cite{chatgpt}, Claude~\cite{claude}, and Gemini~\cite{Anil2023GeminiAF} which are only accessible through APIs. The defining feature of these models is the inaccessibility of their training corpora to researchers, making it challenging to investigate data contamination. Consequently, many recent studies have focused on developing methods to address this issue~\cite{golchin2023time,deng2023investigating}.

\section{Impacts of contamination}
\label{mem_effect}
The contamination effect refers to the extent to which a model exposed to contaminated data during its training phase is influenced by this data in its performance on downstream tasks. Research in this area typically involves selecting a base model and a fixed pre-training corpus, while varying mixture of contaminated data~\cite{magar2022data,jiang2024investigating}. These approaches allow for observing how changes in the contaminated data mix affect downstream task performance. Contrary to intuition, the influence of contamination is complex, as detailed in the following discussions.
\begin{table*}[t]
    \centering
    \small
    \begin{tabular}{llcccc}
        \toprule
        \multirow{2}{*}{\textbf{Method}} & \multirow{2}{*}{\textbf{Level}} & \multirow{2}{*}{\shortstack{\textbf{Access to Training} \\ \textbf{Corpora Required?}}} & \multirow{2}{*}{\shortstack{\textbf{Logits Prob.} \\ \textbf{Required?}}} & \multirow{2}{*}{\textbf{Retrieval?}} & \multirow{2}{*}{\shortstack{\textbf{Prompt-} \\ \textbf{based?}}} \\
        \\
         \cmidrule(lr){1-1}  \cmidrule(lr){2-2}  \cmidrule(lr){3-3}  \cmidrule(lr){4-4}  \cmidrule(lr){5-5} \cmidrule(lr){6-6}
         \citet{brown2020language} & Instance & \cmark & \xmark & \cmark & \xmark  \\ 
          \citet{chowdhery2022palm}   & Instance & \cmark & \xmark & \cmark & \xmark  \\ 
          \citet{touvron2023llama}   & Instance & \cmark & \xmark & \cmark & \xmark  \\ 
          \citet{yeom2018privacy}   & Instance  & \xmark & \cmark & \xmark & \xmark  \\
          \citet{carlini2021extracting}   & Instance  & \xmark & \cmark & \xmark & \xmark  \\
          \citet{dodge2021documenting}   & Instance & \cmark & \xmark & \cmark & \xmark  \\ 
          \citet{carlini2022membership}   & Instance  & \xmark & \cmark & \xmark & \xmark  \\
          \citet{elazar2023whats}    & Instance  & \cmark & \xmark & \cmark & \xmark  \\ 
          \citet{li2023estimating}   & Dataset& \xmark & \cmark & \xmark & \xmark  \\
          \citet{shi2023detecting}   & Dataset& \xmark & \cmark & \xmark & \xmark  \\ 
          \citet{aiyappa2023trust}    & Instance & \xmark & \xmark & \xmark & \xmark  \\ 
          \citet{roberts2023data}   & Instance  & \xmark & \xmark & \xmark & \xmark  \\ 
          \citet{Golchin2023DataCQ}    & Dataset& \xmark & \xmark & \xmark & \cmark \\
          \citet{golchin2023time}   & Both & \xmark & \xmark & \xmark & \cmark \\ 
          \citet{oren2023proving}    & Dataset& \xmark & \cmark & \xmark & \xmark  \\ 
          \citet{deng2023investigating}   & Instance  & \xmark & \xmark & \xmark & \cmark  \\ 
          \citet{bordt2024elephants}    & Instance  & \xmark & \xmark & \xmark & \cmark  \\ 
          \citet{Wei2023SkyworkAM}   & Instance  & \xmark & \xmark & \xmark & \xmark  \\
          \citet{mattern2023membership}   & Instance  & \xmark & \cmark & \xmark & \xmark  \\
          \citet{xu2024benchmarking}   & Instance& \xmark & \xmark & \xmark & \cmark  \\
          \noalign{\vskip 0ex}\noalign{\vskip 0ex} 
         \bottomrule
    \end{tabular}
    
    \caption{Comparison of current data contamination detection method.}
    \label{tab:dataset_comparison}
\end{table*}
\subsection{Memorization and Recall}
Additionally, this area of research is often connected with evaluating the models' ability to memorize information and recall their parametric knowledge~\cite{geva2021transformer,geva2023dissecting,haviv2023understanding,srivastava2023imitation,hartmann2023sok}. Detecting data contamination can be viewed as recalling memorized information to compare it with benchmark data. ~\citet{geva2021transformer} first proposed that feedforward layers in transformers act as key-value memory to store textual patterns from training examples. \citet{haviv2023understanding} identified criteria to trigger memorized knowledge within language models using idioms. Also, ~\citet{hartmann2023sok} categorizes different types of memorization, ranging from low-level linguistic components such as verbatim text to high-level abstractions like alignment algorithms. These methods establish a solid foundation for investigating the impact of data contamination on LLMs. 
\subsection{Task-Level Contamination}
Task-level study of contamination typically involves selecting a specific task, such as classification and question answering. By establishing a fixed benchmark, the extent of data contamination is varied to observe changes in performance. For example,~\citet{magar2022data} pre-train a BERT-based model on a combined corpus of Wikipedia and labeled data from downstream tasks. The findings reveal that while models can memorize data during pre-training, they do not consistently utilize this memorized information in an effective manner. Additionally, the extent of exploitation is affected by several factors, including the duplication of contaminated data and the model size.
\citet{jiang2024investigating} explore the contamination effect of the \emph{decoder-only} architecture using GPT-2. Specifically, they pre-trained GPT-2 on a selected portion of the Pile~\cite{gao2020pile} corpora, intentionally introducing contaminated data during the pre-training phase to assess its impact. Their findings reveal that traditional n-gram-based methods are limited in detecting contamination, and increasing the repetition of contaminated data inversely affects model performance, leading to a performance drop.
\citet{zhu2024critical} also investigate the relation between memorization and generation in the context of critical data size with the configuration of grokking~\cite{power2022grokking}, a phenomenon where a model suddenly achieves near-perfect performance on a task after a period of apparent stagnation during training. 
The authors introduce the Data Efficiency Hypothesis, which outlines three stages of data interaction during model training: insufficiency, sufficiency, and surplus. The study observes that as models grow, they require larger datasets to reach a smooth phase transition.

\subsection{Cross-lingual Contamination}
Most research on task-level contamination is conducted in English. However, in addition to task-level contamination, \citet{Blevins2022LanguageCH} also explore cross-lingual contamination, which refers to when the models are tested for their cross-lingual abilities. For example, pre-training corpora often contain significant amounts of non-English text. If a model is trained on these corpora and then tested on, for example, a Chinese benchmark, the setting is no longer testing the pure cross-lingual generalization, as the model has already been exposed to Chinese text during training. Their research indicates that the corpora utilized for pre-training these models include a significant amount of non-English text, albeit less than 1\% of the total dataset. This seemingly small percentage equates to hundreds of millions of foreign language tokens in large datasets. The study further reveals that these minor proportions of non-English data considerably enhance the models' capability for cross-language knowledge transfer. There is a direct correlation between the models' performance in target languages and the volume of training data available in those languages. In general, more work is required to better understand the impact of contamination in cross-lingual settings. For example, many benchmarks in one target language are direct translations from a benchmark in another source language, and contamination may still occur if models have been exposed to similar contexts in the source language, even though the target evaluation is in another language.

\section{Detecting Data Contamination}
In this section, we discuss various methods for detecting data contamination. We begin with the traditional retrieval-based methods, which are the most straightforward approaches to searching the training data for instances of evaluation data. Particularly, such methods mostly employ n-gram tokenization and string-matching for detection. This approach is also often documented in technical reports of proprietary methods. Subsequently, we introduce several modern methods predominantly developed by the academic community. These methods typically detect contamination indirectly and implicitly, without requiring full access to the training corpora.
\label{detect}
\subsection{Retrieval}
One straightforward approach to detecting contamination is searching the training data for examples that appear in a benchmark. 
This line of research can be approached from two perspectives: the perspective of model developers and that of the academic community.

\subsubsection{Approaches by Model Developers} \label{sec:n-gram}
GPT-3~\cite{brown2020language} was among the first LLMs that incorporated a detailed approach to detecting data contamination in LLMs. The methodology involved filtering the initial training set to eliminate any text from the benchmarks that appeared in the training data. This was achieved by identifying overlaps through searching for 13-gram matches between the test/development sets and the training data.
Overlaps were analyzed using a variable word count, determined by the 5th percentile of example length in words, with a set minimum threshold of 8 words for non-synthetic tasks and a maximum of 13 words for all tasks.

Following this work, LLaMA-2~\cite{touvron2023llama2} employs a similar technique to detect data contamination, combining retrieval methods with n-gram-based tokenization. Specifically, any token n-gram match exceeding 10 tokens indicates contamination. This method facilitates a nuanced analysis of contamination levels, classifying samples as \emph{clean} (\ie, less than 20\% contamination), \emph{not clean} (\ie, 20-80\% contamination), and \emph{dirty} (\ie, more than 80\% contamination). It uses skip-grams longer than 10 tokens and suffix arrays for efficient identification, employing parallel processing to improve speed and scalability. To maintain the integrity of evaluations, Gemini~\cite{geminiteam2024gemini} retrieves and removes any evaluation data that may have been in their training phase.

\subsubsection{Approaches by the Research Community}
Beyond technical reports from model developers, many recent studies by the research community focus on contamination in open-source pre-training corpora commonly used to develop LLMs. This body of research typically involves constructing effective and convenient tools and strategies, developing indexing systems for retrieval, and designing algorithms to determine potential contamination between retrieved passages and benchmark data.
\paragraph{Searching Tools}
To explore different pre-trained corpora, various specialized tools have been developed.  \citet{piktus2023roots} introduce a search engine that spans the entirety of the ROOTS corpus~\cite{laurençon2023bigscience}, featuring both fuzzy and exact search capabilities. Furthermore, \citet{piktus-etal-2023-gaia} present Gaia, a search engine designed based on established principles, providing access to four widely recognized large-scale text collections: C4~\cite{raffel2023exploring}, The Pile~\cite{gao2020pile}, LAION~\cite{schuhmann2022laion5b}, and ROOTS~\cite{laurençon2023bigscience}. Additionally, \citet{elazar2023whats} develop WIMBD, a platform offering 16 analytical tools that enable users to uncover and contrast the contents of vast text corpora.

\paragraph{Indexing System}
The primary limitation of search tools is their dependency on extensive computational resources, combined with the absence of APIs for scalable integration. For individuals endeavoring to develop a custom information retrieval system, \citet{lin2021pyserini} introduce Pyserini, a user-friendly Python-based general-purpose toolkit designed for reproducible information retrieval (IR). Pyserini facilitates various retrieval methods, including sparse retrieval using BM25 with bag-of-words representations, dense retrieval via nearest-neighbor search in transformer-encoded spaces, and a hybrid approach that combines both methods. Researchers also have used such indexing tools to investigate data contamination~\cite{deng2023investigating} for investigating contamination in commonly used pre-training corpora such as The Pile and C4.

\paragraph{Benchmark Overlap Analysis}
To address the challenges of data contamination in language models, \citet{dodge2021documenting} conduct one of the first comprehensive analyses of data contamination by investigating one of the commonly used pre-training corpora, the C4~\cite{raffel2023exploring} 
and they performed analysis on overlaps with the downstream evaluation tasks. This study uncovers a significant volume of text from unexpected sources, including patents and US military websites.
Building on this, \citet{elazar2023whats} presents another comprehensive analysis that explores the overlap between pre-training corpora and the SuperGLUE~\cite{sarlin2020superglue} benchmark. Their findings show significant contamination of several widely used pre-training corpora such as RedPajama \cite{together2023redpajama}, Oscar \cite{li2020oscar}, Pile \cite{gao2020pile} and C4 \cite{raffel2023exploring}, instances of contamination in some SuperGLUE datasets reaching as high as 100\%.

\subsection{Temporal Cutoff}
The concept of time-cutoff implies a significant distinction between models developed or the use of training data up to a certain time point. For instance, GPT-3 was trained using data available only up to September 2021~\cite{chatgpt}. This approach assumes that substantial changes in the dataset's distributions or variances stemming from a specific time cut-off are critically important.

\citet{roberts2023data} conduct one of the first comprehensive longitudinal analyses of data contamination in LLMs. Specifically, they leverage the natural experiment provided by the training cutoffs in GPT models to examine benchmarks released over time. They analyze two code/mathematical problem-solving datasets. Their findings reveal statistically significant trends between LLM pass rates, GitHub popularity, and release dates, which strongly indicate contamination. ~\citet{aiyappa2023trust} also conduct similar experiments to assess performance differences in models before and after their release. Besides, \citet{shi2023detecting} creates a benchmark termed WIKIMIA utilizing data compiled both before and after model training to facilitate accurate detection. Similarly,  \citet{li2023latesteval} employs the most recent data after the time cutoff to develop a benchmark that ensures the newest data, enabling a fair evaluation without contamination.

The time-cutoff technique requires verification that data before and after a specific time-cutoff exhibit distinct distributions with minimal overlap. Additionally, new events or messages extracted from the internet may also overlap with previous ones. For employing a time-cutoff strategy, it is essential to account for and evaluate these potential overlaps in experimental setups.

\subsection{Masking-based}
If a model can accurately predict very specific missing or masked parts of a sentence or paragraph, it likely indicates that the model has previously seen those exact examples during training. Based on this intuition, another line of work in detecting data contamination involves masking-based methods, which mask a phrase or sentence and provide the LLMs with context from a benchmark to guide them in filling in the missing portions. The advantage of this approach is its simplicity and effectiveness, and it doesn't require access to the training data. Below we discuss some representative approaches in this category.

\paragraph{Book-Level}
\citet{chang2023speak} propose the \emph{name cloze} task, wherein names within a book are masked, prompting LLMs to predict the omitted names. This task is specifically designed to evaluate the extent to which models like ChatGPT and GPT-4 have internalized copyrighted content, linking memorization levels to the prevalence of book excerpts online. The findings reveal a notable performance disparity between GPT-4 and ChatGPT in executing the name cloze task, suggesting variations in their capacity to recall and utilize memorized information.

\paragraph{Benchmark-level}
\citet{deng2023investigating} introduce TS-Guessing, a masking-based method designed for benchmark formats to detect data contamination. This technique involves masking an incorrect answer in a multiple-choice question and prompting the model to complete the missing information. It also entails hiding an unlikely word in an evaluation example and requesting the model to generate it. Their findings reveal that several proprietary LLMs can precisely recall the masked incorrect choice in the benchmarks, highlighting a significant potential for contamination in these benchmarks that warrants attention. However, they note that their method depends on the proficient instruction-following capabilities of LLMs. For less capable LLMs, there is a tendency to replicate other choices or produce the correct answer without adhering to the given instructions. 
Part of the work by \citet{xu2024benchmarking} also employs similar methods. Given a sequence, they progressively move forward from the first token and guide LLMs to predict the missing portions of the following part. Their method could be treated as a more quantitative version of \citet{deng2023investigating}, which calculates the results primarily on open-sourced LLMs.

\subsection{Perturbation-based}
Perturbation-based methods involve using various techniques to artificially modify or alter test set samples. This is done to assess if LLMs are overfitting to particular benchmark formats or examples. The objective of this task is to examine whether there is a significant drop or change in performance after applying specific perturbations.

\paragraph{Rephrasing Test Set}
\citet{yang2023rethinking} demonstrate that applying minor alterations to test data, such as rephrasing or translating, can bypass previous n-gram-based detection methods (\S \ref{sec:n-gram}). They reveal that if test data variability isn't eliminated, a 13B model can mimic the performance of state-of-the-art models like GPT-4 by overfitting to benchmarks, as evidenced by their experiments with notable datasets including MMLU~\cite{hendrycks2021measuring}, GSM8K~\cite{cobbe2021training}, and HumanEval~\cite{chen2021evaluating}. To address this growing issue, they propose a new LLM-based detection approach using embedding similarity search to identify the top-k training samples most similar to each test sample. It then uses a high-quality language model (like GPT-4) to evaluate whether any of those top-k samples are semantically too close to the test sample.
In a recent paper, ~\citet{dekoninck2024constat} proposed ConStat, a novel method for detecting and quantifying contamination in LLMs. The authors redefine contamination from a performance-based perspective, considering it as an artificially inflated benchmark performance that fails to generalize to real-world tasks. ConStat employs a statistical test that compares a model's performance on the original benchmark to its performance on carefully selected reference benchmarks while accounting for differences in difficulty using a set of uncontaminated reference models. 

\paragraph{Creating Reference Set}
In addition to directly rephrasing test set examples, \citet{Wei2023SkyworkAM} use GPT-4 to create a reference set resembling the test set. They then calculate the difference between the reference set and the test set to assess the contamination issues potentially caused by intentional data contamination. Higher differences indicate a greater potential for data leakage.

\subsection{Canonical order}
The canonical assumption posits that if a model has been exposed to data from a dataset, it will exhibit a preference for the canonical order provided by the dataset from public repositories, as opposed to datasets that have been randomly shuffled.

\citet{oren2023proving} develop a sensitivity test to detect biases in the canonical order of benchmark datasets used for LLMs. Based on the principle that, in the absence of data contamination, any permutation of an exchangeable benchmark dataset should be equally likely, they create a methodology capable of identifying contamination through the model's preference for specific data orderings. Their method can identify biases in models with as few as 1.4B parameters using only 1,000 test examples and remains effective even with minimal dataset representation. However, the approach has limitations: if the model preprocesses or shuffles the benchmark data during pre-training, detecting contamination based on order becomes difficult.

\subsection{Behavior Manipulation}
We term behavior observation as a new perspective that leverages different perspectives of controlling experiments related to the test set. This is done by observing whether the behavior (\ie, output and selection choice) are different.

\citet{golchin2023time} propose a dual-layered approach for identifying contamination in LLMs at both the instance and partition levels. The initial phase employs \emph{guided instruction}, a technique that utilizes a specific prompt incorporating the dataset name, partition type, and an initial segment of a reference instance. This prompt encourages the LLM to generate a completion. An instance is considered contaminated if the LLMs' output closely resembles or exactly matches the subsequent segment of the reference. Building on this concept, \citet{Golchin2023DataCQ} introduces a novel methodology by devising a data contamination quiz. This quiz presents a set of choices, including one from the test set and others that are variations of the original instance. The model is then tasked with selecting an option, and its decision is used to assess contamination based on its choice. This approach not only follows the general pattern of contamination detection but also offers a unique perspective by varying the format of the choices provided to the model. 

Besides,~\citet{dong2024generalization} propose CDD (Contamination Detection via Output Distribution) for detecting data contamination and TED (Trustworthy Evaluation via Output Distribution) for mitigating its impact on evaluation. CDD identifies contamination by analyzing the peakedness of the LLM's output distribution using only the sampled texts, while TED corrects the output distribution to ensure trustworthy evaluation.

To employ methods based on this assumption, researchers must verify that behavior differences are solely attributable to data contamination, particularly in contrast to variations arising from random prompt perturbation.

\subsection{Membership Inference Attacks}
Membership Inference Attacks (MIA) aim to determine whether a specific data point was used in the training data of a target model. While MIA is a well-established concept in traditional machine learning~\cite{shokri2017membership, hu2022membership}, their application in the context of LLMs has been relatively understudied. 
This subsection explores the application of MIA to LLMs, demonstrating their utility in detecting contamination.
\paragraph{Background}
\citet{yeom2018privacy} measure the perplexity of a sample to measure the memorization of training data. \citet{carlini2021extracting} build upon this work to further improve precision and reduce the false negative rate by considering the intrinsic complexity of the target point. Furthermore, \citet{carlini2022membership} calibrate the sample’s loss under the target model using the sample’s zlib compression size.

\paragraph{Applying MIA to LLMs}
\citet{mattern2023membership} introduce and assess neighborhood attacks as a novel method to evaluate model vulnerabilities without requiring access to the training data distribution. They use an estimate of the curvature of the loss function at a given sample, which is computed by perturbing the target sequence to create $n$ neighboring points and comparing the loss of the target $x$ with its neighbors. By comparing model scores of a given sample with those of synthetically generated neighbor texts, this approach seeks to understand if model fragility can enhance security. 

Recently, \citet{shi2023detecting} introduces MIN-K\%, a method that utilizes the $k\%$ of tokens with the lowest likelihoods to compute a score, rather than averaging over all token probabilities as in traditional loss calculations. This approach is based on the hypothesis that an unseen example is likely to contain a few outlier words with low probabilities under LLMs, whereas a seen example is less likely to feature words with such low probabilities. Additionally,~\citet{ravaut2024llms} also introduces LLMSanitize, a library to implement contamination detection methods.

Additionally,~\citet{ye2024data} propose Polarized Augment Calibration (PAC), a novel approach for detecting training data contamination in black-box LLMs. PAC extends the MIA framework by leveraging confidence discrepancies across spatial data distributions and considering both distant and proximal probability regions to refine confidence metrics.~\citet{duarte2024decop} use DE-COP to probe language models with multiple-choice questions to determine whether LMs memorize certain parts of books.

MIA, in the context of LLMs, is typically based on perplexity or variations derived from language model perplexity. This implies reliance on the output logit probability from the language models. However, its statistical simplicity also offers significant advantages compared to other detection methods that require careful validation of assumption~\cite{duan2024membership}.

\section{Mitigating Data Contamination}
\label{mitigating}
Without specific mitigation strategies, the continuous development of new benchmarks—often released publicly on the internet— and deprecating the old benchmarks does \textit{not} resolve contamination issues, as newer models can access this data. Consequently, several studies have proposed mitigation approaches to address this problem. In this section, we will introduce these strategies from the perspectives of benchmark construction, updating, encryption, and protection.
\paragraph{Benchmark Construct Selection}
\citet{li2023latesteval} propose to construct evaluation benchmarks from the most recent texts, thus minimizing the risk of overlap with the pre-training corpora. 

\paragraph{Dynamic Benchmark Refreshing}
\citet{Zhu2023DyValGD} introduce a dynamic evaluation protocol that utilizes directed acyclic graphs to generate evaluation samples of varying complexities, aiming to address the static and potentially contaminated nature of existing benchmarks. Besides, \citet{zhu2023cleaneval} provide Clean-Eval, which utilizes LLMs to paraphrase and back-translate contaminated data, creating a set of expressions that convey the same meaning in varied forms. This process generates a candidate set from which low-quality samples are filtered out using a semantic detector. The final selection of the best candidate from this refined set is based on the BLEURT~\cite{sellam2020bleurt} score, ensuring the chosen expression is semantically similar to the original data but articulated differently. Furthermore, ~\citet{Zhou2023DontMY} also suggest providing a diverse set of prompts for testing, which offers a dynamic evaluation to mitigate data contamination.

\paragraph{Benchmark Data Encryption}
~\citet{Jacovi2023StopUT} suggests that test data released to the public should be safeguarded through encryption using a public key, and the distribution of derivatives should be strictly prohibited by the licensing agreement. To implement this, the recommended approach is toencrypt the test data before uploading it. This can be efficiently done by compressing the data into a password-secured archive.

\paragraph{Benchmark Label Protection} 
\citet{Jacovi2023StopUT} and \citet{Zhou2023DontMY} emphasize the critical need to safeguard the ground truth labels of test datasets. These labels can inadvertently be exploited during the training phase, or even intentionally after being rephrased. Providing both the question and its context is an effective strategy to prevent such deliberate contamination.

\section{Discussion and Future Directions}
This section will address the impact, detection, and mitigation of previously introduced data contamination and explore the topic at a higher level. We aim to offer more insights into the current challenges, necessity, and robustness of detecting data contamination methods. We will also discuss how these concepts can be applied in more realistic settings. Additionally, we will consider data contamination as an overarching research direction and explore potential future pathways for this field.
\paragraph{Challenges for Detecting Black-Box Models}
The primary challenge in evaluating different methods for detecting data contamination in large language models is the absence of a ground truth label, \ie, a benchmark dataset comprising entirely contaminated data. This absence creates difficulties in comparing the effectiveness of various detection techniques designed for black-box models. One alternative approach involves fine-tuning the model using test set labels to create artificially contaminated data. However, the question remains whether the scenarios of contamination during the pre-training phase and the fine-tuning phase are consistent. Additionally, due to limited access to the complete training corpus, we can only generate fully contaminated data, making it challenging to obtain fully uncontaminated data. This situation complicates efforts to accurately assess and compare the efficacy of contamination detection methods.

\paragraph{Evading Detection of Data Contamination}
\citet{dekoninck2024evading} highlights the ease with which MIA detection methods can be evaded. These methods, some of which are also employed for identifying data contamination, have been criticized in prior research. Notably, the efficacy of n-gram-based substring detection is questioned due to its numerous vulnerabilities and susceptibility to manipulation~\cite{Zhou2023DontMY,deng2023investigating,jiang2024investigating}. Beyond the traditional n-gram and MIA approaches, recent studies have demonstrated that several contemporary techniques can be compromised through targeted attacks. For instance, by integrating a dataset with a significantly large pre-trained dataset, one can disrupt the canonical order assumption, thereby undermining its integrity. 

\paragraph{From Memorization to Exploitation}
Drawing a definitive conclusion about the correlation between memorization and exploitation (\ie, performance on downstream tasks) remains challenging. Various factors can impact the outcomes observed in our study, including differences in model architecture, the repetition of contaminated data, the strategies employed during pre-training or fine-tuning phases, and the training principles used like RLHF+PPO~\cite{zheng2023secrets} and DPO~\cite{rafailov2023direct}. These elements can significantly influence the models' downstream task performance.

\paragraph{Detecting or Mitigating?}
Currently, there is an increasing focus on developing novel methods for detecting data contamination, which is crucial for investigating and understanding data contamination scenarios. Effective detection tools can also help prevent intentional data contamination to a certain extent. However, there remains a significant need for research focused on mitigating data contamination. The research question arises: how can we create a dynamic evaluation method that uses potentially contaminated benchmarks to provide clean evaluations? In recent developments, many have started leveraging language models as agents to perform various tasks. An intriguing future direction could be to utilize LLMs as 'Benchmark Agents' to offer various forms of evaluation that convey the same meaning. 
\paragraph{How to Create Benchmarks without Data Contamination}
To address the challenge of creating a benchmark free from data contamination, it is essential to consider innovative approaches. Firstly, an effective strategy involves constructing a dataset significantly larger than the target size, which can be in future refined. The excessive size allows for the application of rigorous data contamination checks to refine the dataset down to the initial target size. Additionally, the implementation of a unified, reliable, and dynamic evaluation framework is crucial. Such a framework offers the flexibility to adaptively assess benchmarks across various formats, enhancing the robustness of the evaluation process. Beyond these broader strategies, a practical yet profound method involves generating content that is rare or virtually nonexistent on the Internet or other public domains. 
\paragraph{Rethinking Evaluation Paradigms} As the scale of models and training datasets expands, it's conceivable that in future, the majority of tasks of interest or practical relevance might already fall within the distribution covered during training of LLMs. If in that hypothetical world, future models can execute all tasks that we care about with high accuracy, the relevance of whether these exact tasks were encountered during training diminishes. In such a scenario, the traditional emphasis on generalization—distinguishing between training and testing instances—might not be as critical. This raises important considerations about the trade-off between the usefulness of tasks that are well-represented in the data and the ability to generalize to entirely new scenarios. Consequently, this could necessitate a reevaluation of standard machine learning evaluation paradigms. 

\section{Conclusion}
In this paper, we present an extensive survey on the topic of data contamination in large language models. We start by laying the groundwork with a discussion on the effect of contamination, setting the stage for a deeper examination of various data contamination detection methods. We critically analyze the assumptions underlying these methods, highlighting their limitations and the prerequisites for their application. Subsequently, we explore strategies for mitigating data contamination, addressing potential challenges, and suggesting directions for future research in this area. Our goal is to provide a comprehensive guide for researchers seeking a systematic understanding of data contamination. We also aim to underscore the critical importance of this topic, advocating for increased attention due to its pressing relevance.

\section{Limitations}
It is challenging to provide a quantitative comparison between different data contamination detection methods due to their varying assumptions and requirements. Ideally, we would conduct a quantitative analysis to assess the effectiveness of these methods, assigning rankings or benchmarks to discuss their advantages and disadvantages. Another limitation of the survey paper is the difficulty in categorizing each method into a single, definitive class. For instance, \citet{shi2023detecting} not only offers benchmarks and analyses but also proposes a detection method. Similarly, \citet{Zhou2023DontMY} discusses both the detection of contamination and strategies for its mitigation. Our approach primarily classifies each work into its most evident category.

\section{Ethics Statement} 
In our survey paper, which examines the impact of data contamination alongside methods for its detection and mitigation, we assert that our work not only adheres to ethical standards and avoids potential misuse issues but also offers a comprehensive summary that contributes to the fair and transparent evaluation of large language models. This positions it as a valuable resource for promoting fairness and transparency within the community.
\section{Acknowledgement}
This project was supported in part by Tata Sons Private Limited, Tata Consultancy Services Limited, and Titan. We extend our sincere thanks to Simeng Han, Yixin Liu, and Hanjie Chen for their invaluable proofreading and insightful discussions.

\bibliography{anthology,custom}
\bibliographystyle{acl_natbib}

\end{document}